\documentclass[conference]{IEEEtran}
\IEEEoverridecommandlockouts
\usepackage{cite}
\usepackage{amsmath,amssymb,amsfonts}
\usepackage{algorithmic}
\usepackage{graphicx}
\usepackage{textcomp}
\usepackage{xcolor}
\usepackage{indentfirst}
\usepackage[numbers,sort&compress]{natbib}

\def\BibTeX{{\rm B\kern-.05em{\sc i\kern-.025em b}\kern-.08em
    T\kern-.1667em\lower.7ex\hbox{E}\kern-.125emX}}
\begin{document}

\title{Learning Behavior Recognition in Smart Classroom with Multiple Students Based on YOLOv5\\
\thanks{Corresponding author: Zhifeng Wang, Email: zfwang@ccnu.edu.cn}
}

\author{\IEEEauthorblockN{1\textsuperscript{st} Zhifeng Wang}
\IEEEauthorblockA{\textit{CCNU Wollongong Joint Institute} \\
\textit{Central China Normal University}\\
Wuhan 430079, China \\
zfwang@ccnu.edu.cn}
\and
\IEEEauthorblockN{2\textsuperscript{nd} Jialong Yao}
\IEEEauthorblockA{\textit{CCNU Wollongong Joint Institute} \\
\textit{Central China Normal University}\\
Wuhan 430079, China \\
jy123@uowmail.edu.au}
\and
\IEEEauthorblockN{3\textsuperscript{rd} Chunyan Zeng}
\IEEEauthorblockA{\textit{School of Electrical and Electronic Engineering} \\
	\textit{Hubei University of Technology}\\
	Wuhan 430068, China \\
	cyzeng@hbut.edu.cn}
\and
\IEEEauthorblockN{4\textsuperscript{th} Wanxuan Wu}
\IEEEauthorblockA{\textit{CCNU Wollongong Joint Institute} \\
\textit{Central China Normal University}\\
Wuhan 430079, China \\
wuwx0428@163.com}
\and
\IEEEauthorblockN{5\textsuperscript{th} Hongmin Xu}
\IEEEauthorblockA{\textit{Faculty of Artificial Intelligence in Education} \\
\textit{Central China Normal University}\\
Wuhan 430079, China \\
xhm@ccnu.edu.cn}
\and
\IEEEauthorblockN{6\textsuperscript{th} Yang Yang}
\IEEEauthorblockA{\textit{CCNU Wollongong Joint Institute} \\
	\textit{Central China Normal University}\\
	Wuhan 430079, China \\
	univeryang@ccnu.edu.cn}
}
\maketitle

\begin{abstract}
Deep learning-based computer vision technology has grown stronger in recent years, and cross-fertilization using computer vision technology has been a popular direction in recent years. The use of computer vision technology to identify students' learning behavior in the classroom can reduce the workload of traditional teachers in supervising students in the classroom, and ensure greater accuracy and comprehensiveness. However, existing student learning behavior detection systems are unable to track and detect multiple targets precisely, and the accuracy of learning behavior recognition is not high enough to meet the existing needs for the accurate recognition of student behavior in the classroom. To solve this problem, we propose a YOLOv5s network structure based on you only look once (YOLO) algorithm to recognize and analyze students' classroom behavior in this paper. Firstly, the input images taken in the smart classroom are pre-processed. Then, the pre-processed image is fed into the designed YOLOv5 networks to extract deep features through convolutional layers, and the Squeeze-and-Excitation (SE) attention detection mechanism is applied to reduce the weight of background information in the recognition process. Finally, the extracted features are classified by the Feature Pyramid Networks (FPN) and Path Aggregation Network (PAN) structures. Multiple groups of experiments were performed to compare with traditional learning behavior recognition methods to validate the effectiveness of the proposed method. When compared with YOLOv4, the proposed method is able to improve the mAP performance by 11\%.\par
\end{abstract}


\section{Introduction}
With the massive use of cloud classrooms and online classes in recent years, and the smart classroom being a key national development project, the need to record and identify student performance in the classroom is increasing. There has been much research in the education sector to predict and judge students' future performance and achievement in the classroom through their movement and behavioral performance. Traditionally, teachers have maintained discipline in the classroom and manually recorded students' performance in class. This approach is inefficient, does not provide a holistic view of the classroom and can be easily influenced by the subjective nature of the teacher. Nowadays, the use of computer vision technology to identify the status of students in the classroom has become an important technological basis for building smart classrooms \cite{SC_CV}. Deep learning techniques can provide data to support the building of intelligent classrooms, making a statistical judgement on student behavior in the background and enabling better assessment of a student's learning \cite{Lyu2022}. 

The current mainstream target detection algorithms are broadly divided into two categories, one for two-stage type, such algorithms are represented by the Faster R-CNN \cite{RCNN,Zeng2022b,Wang2022ac} series of algorithms, which first extract the pre-selected boxes in the image and then based on this training and classification of the target. This class of methods has a long processing time for images but has high accuracy. The one-stage algorithm, represented by the YOLO series \cite{YO,Wang2021}, single shot multi box detector (SSD) \cite{r8,Zeng2022}, on the other hand, inputs the whole image into the neural network and then outputs the position and category information directly. It is characterised by being faster and more adaptable to the need for real-time recognition of video images in reality.

Inspired by the previous research work, we use YOLOv5 as the main network framework, which can maintain good accuracy while being fast. This paper also introduces the Squeeze-and-Excitation Networks (SENet) attention detection mechanism, which can achieve better accuracy in scenes with complex backgrounds. The YOLOv5 algorithm was used to detect seven patterns of gestures commonly performed by primary and secondary school students in the classroom: looking up to listen, raising hands, looking down to read, looking down to write, standing up, turning head, and group discussion. This paper is compared with a detection framework using the YOLOv4 network structure used in this paper. The contributions of this paper can be summarized as follows:

\begin{figure*}[t]
	\centering
	\includegraphics[width=1\textwidth]{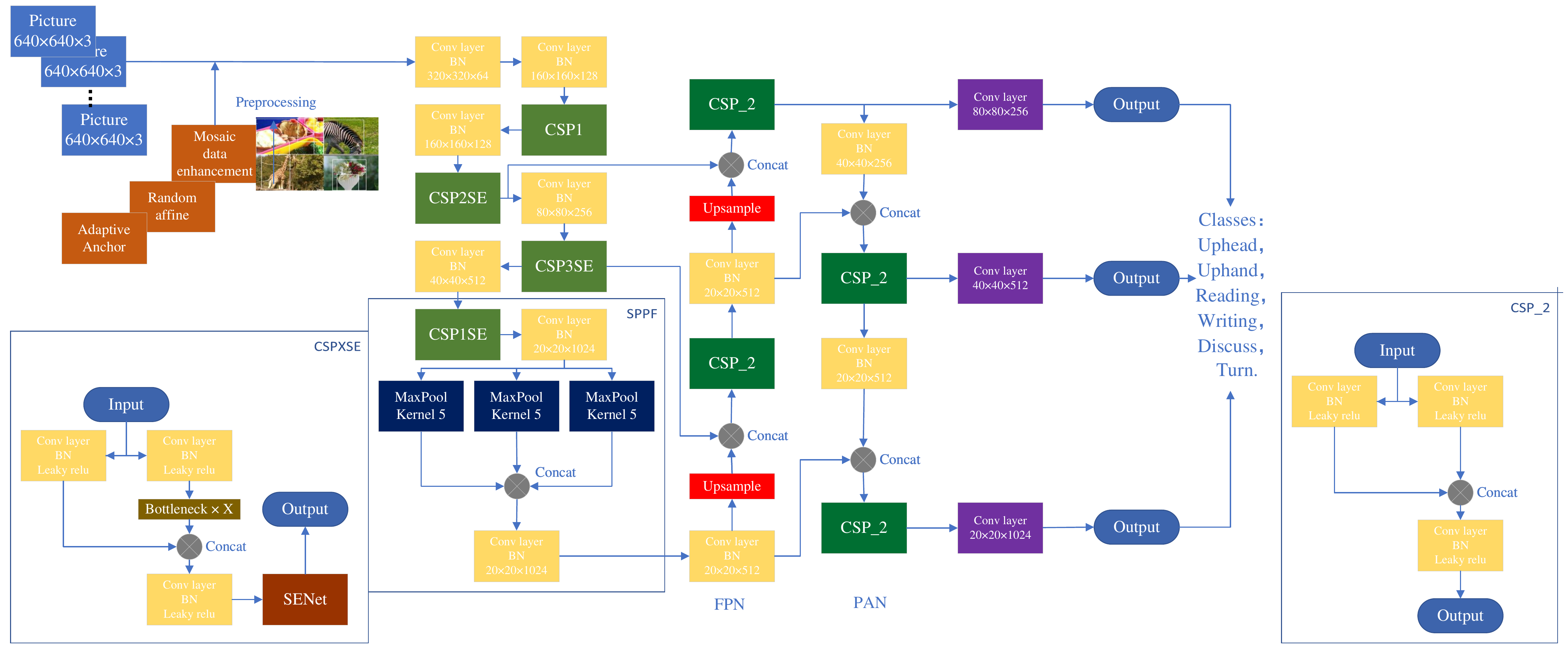}
	\caption{The framework of students learning behavior recognition and analysis. It consists of a head section for pre-processing, a backbone section for extracting features, an SPPF section for fusing features, and finally an FPN and PAN structure section for classification and aggregation.}
	\label{fig:mesh1}
\end{figure*}

\begin{itemize}
	\item In this paper, we propose a YOLOv5 enhanced learning behavior recognition method which can precisely recognize multiple students' learning behavior. What's more, the SE attention detection mechanism is included to enhance the performance of recognition. 
	\item By comparing with YOLOv4, the students' learning behavior recognition network proposed in this paper is able to improve the mAP by 11\% and has a significant advantage in recognising the common classroom behaviors of students turning their heads, raising their hands and listening with their heads up. It is more than capable of meeting the needs of real-time classroom detection.
\end{itemize}

The rest of the paper is structured as follows: in Section 2, student classroom behavior recognition and related research on YOLO are described. The identification process and network framework for student behavior recognition is presented in Section 3, while the comparative and intersectional experiments in this paper are demonstrated in Section 4. Finally, the conclusions of this paper are presented in Section 5.

\section{Related Work}
\subsection{Learning Behavior Recognition}
The behavioral performance of students in the classroom reflects to some extent the quality of the teacher's classroom. Serious expressions and performance mean that the teacher's classroom is of high quality in this segment and is able to engage students in following the teacher's progress and thinking. If, on the other hand, students have negative classroom behavior, this indicates that they are bored with the current class content \cite{lb,Wang2022t}. However, as a lesson progresses, students' attentional performance changes dynamically at different stages of the classroom, and the analysis of student behavior, which in turn helps teachers to evaluate and improve their teaching, is a major direction in the field of smart education. A typical classroom behavioral analysis method, S-T analysis \cite{r10,Zeng2022a}, is used to improve teachers' teaching methods by observing the behavior of both teachers and students separately and using these two behavioral analyses to obtain correlations between them. In addition, Jia Pengyu et al. judged students' listening by looking at their emotion recognition \cite{r11,Zeng2021a}, and Wu Lijuan and Liang Dai Li et al., by using a non-maximal suppression algorithm to optimise the SSD target detection algorithm, were eventually able to recognise two head movements and four postures of students \cite{r15,Zeng2020}.

\subsection{Target detection based on YOLO}
The YOLO algorithm used in this paper was proposed by Joseph Redmon. Unlike previous algorithms that first traverse the image to create several frames to be detected and then perform classification detection, the YOLOv1 algorithm inputs the whole image into the neural network, predicts the image directly, and regresses to calculate the location and category information of the object to be detected in the image. In YOLOv2, Joseph Redmon introduced normalization and anchor mechanisms to improve the accuracy of recognition. In YOLOv3, a Darknet-53 full convolutional residual network is used to downsample the input image seven times and the output is obtained in three different dimensions by using FPN to detect targets of three sizes, large and small. This approach has greatly improved the accuracy of YOLO in detecting small and overlapping targets \cite{r16,Wang2017}. In the YOLO v4 version after fusing the deep information with the low level using the FPN structure, the PAN structure was introduced to feed the low-level information to the deep neural network. This enables a further increase in the detection accuracy of the target. The student classroom behavior recognition system used in this paper adds an attention monitoring mechanism to the network structure of YOLOv5, allowing the network to focus more on recognising information about the object itself and reducing the interference of the background on the recognition results. In addition, data augmentation techniques are introduced to make the recognition scenarios more extensive and the applications more universal.

With the YOLO family of algorithms, Zhu Chao used the Faster R-CNN algorithm to feature recognition of student performance in the classroom, and then used the YOLOv3 algorithm to classify the results \cite{r12}. Huang, Chang-Rong and Qi, Sheng-En proposed an improved YOLOv4 algorithm to detect students' posture in the classroom, using Mosaic data enhancement and K-Means clustering analysis to obtain better recognition results \cite{r13,Wang2018a}. In addition, Wang and Shen used the OpenPose algorithm to identify the global features of the human body and determine whether the human body is in a head-down or head-up state by using the angle of the joints. The YOLO algorithm was also used to capture hand information to classify whether students were playing with their mobile phones \cite{r14}.

\section{Proposed Method}
The YOLO family of algorithms is currently the dominant algorithm in the field of target detection, and as a classical algorithm with one-stage architecture, YOLO combines accuracy and speed and is widely used in various fields. Inspired by the success of deep learning in the fields of data mining \cite{Wang2023,Lyu2022,Li2023a,Min2019,Wang2022as}, computer vision \cite{Wang2021,Zeng2021c,Wang2022ac,Zeng2020,Li2023,Zeng2022,Wang2017,Zeng2022b,Wang2015a,Zeng2020a,Tian2018a,Min2018,Wang2022at} and speech processing \cite{Wang2021m,Zeng2021a,Wang2022t,Zeng2022a,Wang2020h,Zeng2021b,Wang2018a,Zeng2018,Wang2015b,Zhu2013,Wang2011,Wang2011a}, the main network framework proposed in this paper is YOLOv5s, with 45 convolutional layers, 5 kernel maximum pooling layers and output layers consisting of FPN and PAN networks.

\subsection{Student Behavioral Testing Process}\label{AA}
When performing detection, the network will first divide the input image into $S \times S$ grid cells, and each grid will make a judgement that if there is a centroid of an object within the grid, then that grid is responsible for detecting that object. When a grid is ready to detect an object, it outputs $B$ bounding boxes and $C$ information about the probability that the object belongs to a certain group. The bounding box contains five data values, which are the two-dimensional coordinate information of the object $(x, y, w, h)$ and the confidence level. Where $x, y$ represent the centre coordinates of the bounding box obtained from the current grid, $w, h$ represent the width and height of the bounding box. The confidence level reflects whether the current bounding box contains an object and the accuracy of that object:
\begin{equation}
Confidence=pr(object)*IoU
\end{equation}
The value of $Pr(Object)$ depends on whether or not there is an object in the current grid. If there is, the value is 1; if there is no object, the value is 0. The second term is $IoU$ (Intersection over Union), which is used to calculate the degree of overlap between the generated bounding box and the ground truth, and is calculated by the formula:
\begin{equation}
IoU = \frac{BoundingBox~ \cap ~GroundTruth}{BoundingBox~ \cup~GroundTruth}~
\end{equation}\par
After sending the images into the student behavior detection network, the images were preprocessed by convolutional layers. A single image of size $640 \times 640 \times 3$ is segmented into images of size $320 \times 320 \times 64$. This reduces the number of subsequent operations without losing as much information as possible. The image is then downsampled 32 times through a convolutional layer and a CSP structure, and the output is fed into the SPPF module for stitching of the feature information. The CBS structure is then used for feature extraction. Finally, the FPN and PAN structures are used to fuse the deep and shallow feature information, output the detection category and frame the object.

\subsection{Squeeze-and-Excitation Networks attention mechanism}
The SE attention mechanism encodes the entire spatial feature in a channel as a global feature through automatic learning utilizing global averaging pooling by means of the squeeze operation \cite{r17}.\par
\begin{equation}
z_{c} = \frac{1}{H*W}{\sum\limits_{i = 1}^{H}{\sum\limits_{j = 1}^{W}{x_{c}(i,j)}}}
\end{equation}\par
where $z_{c}$ is the output associated with the $c$ channel and the input $x$ is a convolutional layer with a fixed kernel size. An excitation operation is then performed to obtain the importance of each channel in the feature map.\par
\begin{equation}
s = F_{ex}\left( {z,W} \right) = \sigma\left( {g\left( {z,W} \right)} \right) = \sigma\left( W_{2}\delta\left( W_{1}z \right) \right)
\end{equation}\par
$\sigma$ is the sigmoid activation function and $\delta$ is the relu activation function. Different weight values are assigned to each channel according to their different levels of importance. This boosts the proportion of weights for channels that benefit detection accuracy while suppressing the influence of those channels that are not important. After the above steps, the normalised weights obtained earlier are weighted onto the features of each channel by scale.\par

\subsection{Network Design of Student Behavior Detection System}
The network structure of this paper is implemented by improving the network structure of YOLOv5s, which can be divided into four parts: head, backbone, neck and prediction. The head is the input side, and the training side adopts the Random affine data enhancement technique to stitch and rotate the dataset images on the fly and the Mosaic data enhancement technique to randomly select four images, randomly scale them and stitch them on the fly, this operation not only greatly enhances the breadth of the dataset. It also solves the problem that the distribution of small target datasets is not uniform enough during training. As well as the adaptive anchor calculation method can use different sizes of anchors for different bounding boxes, enabling the network to better learn the feature information of the targets.
In the backbone section, the CSP and SPPF structures are used. the CBS structure consists of a convolutional layer and Batch normalization with a SiLu activation function at the end.
\begin{equation}
SiLu(x)=x*sigmoid(x)
\end{equation}\par
 the CSP structure consists of two CBS modules as and a Bottleneck, the output of the Bottleneck is con conact splicing. After obtaining deep target features through this residual network, feature extraction is performed through a layer of the CSP structure, followed by the use of the SENet attention detection mechanism to reduce feature extraction from different backgrounds for the same detection target
\begin{figure}[h]
    \centering
    \includegraphics[width=0.25\textwidth]{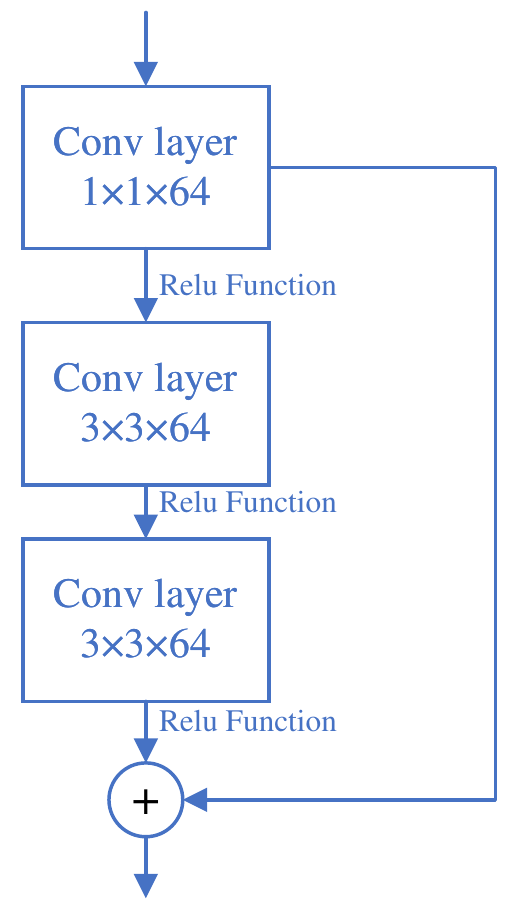}
    \caption{Bottleneck structure.}
    \label{fig:mesh1}
\end{figure}
After 32-fold downsampling of the backbone network with 5 convolutional layers in steps of 2, the output is fed to the SPPF module to stitch the target features extracted by the neural network through 3 maximum pooling layers with a kernel of 5, followed by a CBS for feature extraction.
And in the Neck part, the structure of FPN and PAN was adopted. The idea of using multiple scales in the FPN structure to detect targets of different sizes \cite{r18} allows the output of three feature maps with sizes of 20×20, 40×40 and 80×80. The finer the scale the higher the accuracy of the feature maps in recognising small objects, which makes it possible to have a good accuracy rate in recognising these small objects as well. The FPN structure passes the higher-level feature information to the lower levels by up-sampling to obtain a predictive feature map. The upward FPN structure is followed by a top-down downsampled FPN network with a PAN structure added to it. Thus strong semantic features are passed top-down using the FPN layer, while the PAN passes strong localisation features bottom-up.

\subsection{Loss Function}
YOLOv5s uses CIoU-Loss as the loss function \cite{r19}, and three important geometric indicators of the Bounding box: IoU, the distance between the four endpoints and the centroid, and the aspect ratio as factors for judging the Bounding box regression function.
\begin{equation}
CIoU = IoU - \frac{\rho^{2}\left( {b,b^{gt}} \right)}{c^{2}} - \alpha v
\end{equation}
where $\rho^{2}\left( {b,b^{gt}} \right)$ denotes the Euclidean distance between the Bounding box and the centroid of the ground truth. $c$ denotes the diagonal distance representing the smallest closed region that can contain both the prediction box and the ground truth. $alpha$ and $v$ are given by the following equations, where $w^{gt}$ and $h^{gt}$ denote the width and height of the ground truth, respectively.
\begin{equation}
\alpha = \frac{v}{1 - IoU + v}\\
\end{equation}
\begin{equation}
v = \frac{4}{\pi^{2}}\left( arctan\frac{w^{gt}}{h^{gt}} - arctan\frac{w}{h} \right)^{2}
\end{equation}
In the final stage of detection, NMS (Non-maximum Suppression) is performed on all the bounding boxes to filter out the ones with the highest overlap with the ground truth.

\section{Experimental Results and Analysis}
\subsection{Experimental setup environment}\label{AA}
The experiment was compiled and tested using python 3.8, the corresponding development tool was VSCode 2019, the main computer vision library was python-OpenCV 4.4.0, the deep learning framework used was Pytorch v1.12.1, the operating system was Windows 10, and the CPU was a hexacore processor with i7-10750H The main frequency is 2.6GHz, the graphics card is RTX2070-MaxQ, the RAM is 16GB, and the hard disk capacity is 1TB SSD.

\subsection{Data set collection and training}
The source of the dataset was the classroom video recordings of the 2019 Primary School Language Grades 1-6 Departmental Excellence Classes on the China National Resources Public Service Platform obtained by taking screenshots at one-second intervals, with a frame rate of 25 f/s. A total of 8,884 images were taken using the lambelme software for the seven states of uphead, uphand, reading, writing, stand, turn and discuss. The corresponding json files were obtained. The corresponding JSON files were obtained, in which the category of the target to be detected and its position in the picture was recorded.

\begin{figure}[h]
    \centering
    \includegraphics[width=0.5\textwidth]{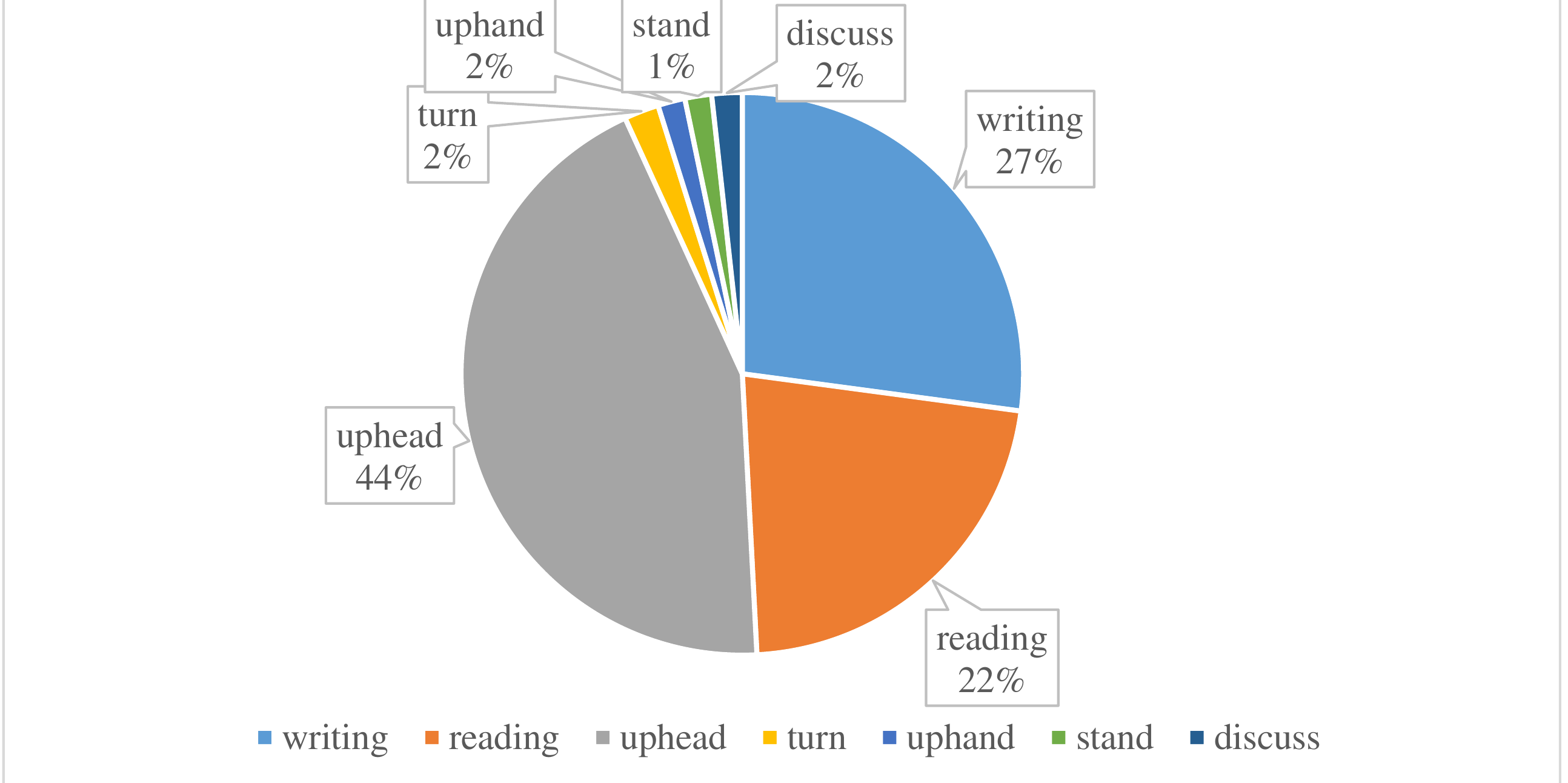}
    \caption{Student behavior dataset annotation distribution.}
    \label{fig:mesh1}
\end{figure}

\subsection{Analysis of experimental results}
Depending on whether the target is correctly identified, the correct result is named TP (True Positive), the incorrect result is called FP (False Positive) and the correct target is missed is named FN (False Negative). From the above parameters, two criteria can be used to judge the results of the experiment, Precision and Recall: 
\begin{equation}
    \centering
\begin{aligned}
precision~ = ~\frac{TP}{TP + FP}\\
recall~ = ~\frac{TP}{TP + FN}
\end{aligned}
\end{equation}

\begin{table}[]
    \centering
\begin{tabular}{ccc}
\multicolumn{3}{c}{Table 1. The performance of YOLOv5s SE.}                                 \\ \hline
\multicolumn{1}{l}{epoch} & precision & recall \\ \hline
0                         & 0.7562            & 0.13094        \\
10                        & 0.70842           & 0.6163         \\
20                        & 0.8042            & 0.71241        \\
30                        & 0.83339           & 0.74744        \\
40                        & 0.83301           & 0.76378        \\
50                        & 0.83717           & 0.78106        \\
60                        & 0.84247           & 0.78817        \\
70                        & 0.85659           & 0.79099        \\
80                        & 0.84363           & 0.79851        \\
90                        & 0.84962           & 0.80459        \\
100                       & 0.85142           & 0.8104         \\
110                       & 0.86556           & 0.79678        \\
120                       & 0.85363           & 0.80995        \\
130                       & 0.86134           & 0.80878        \\
140                       & 0.86247           & 0.81084        \\
150                       & 0.86163           & 0.81417        \\ \hline
\end{tabular}
\end{table}
\begin{figure}[h]
    \centering
    \includegraphics[width=0.5\textwidth]{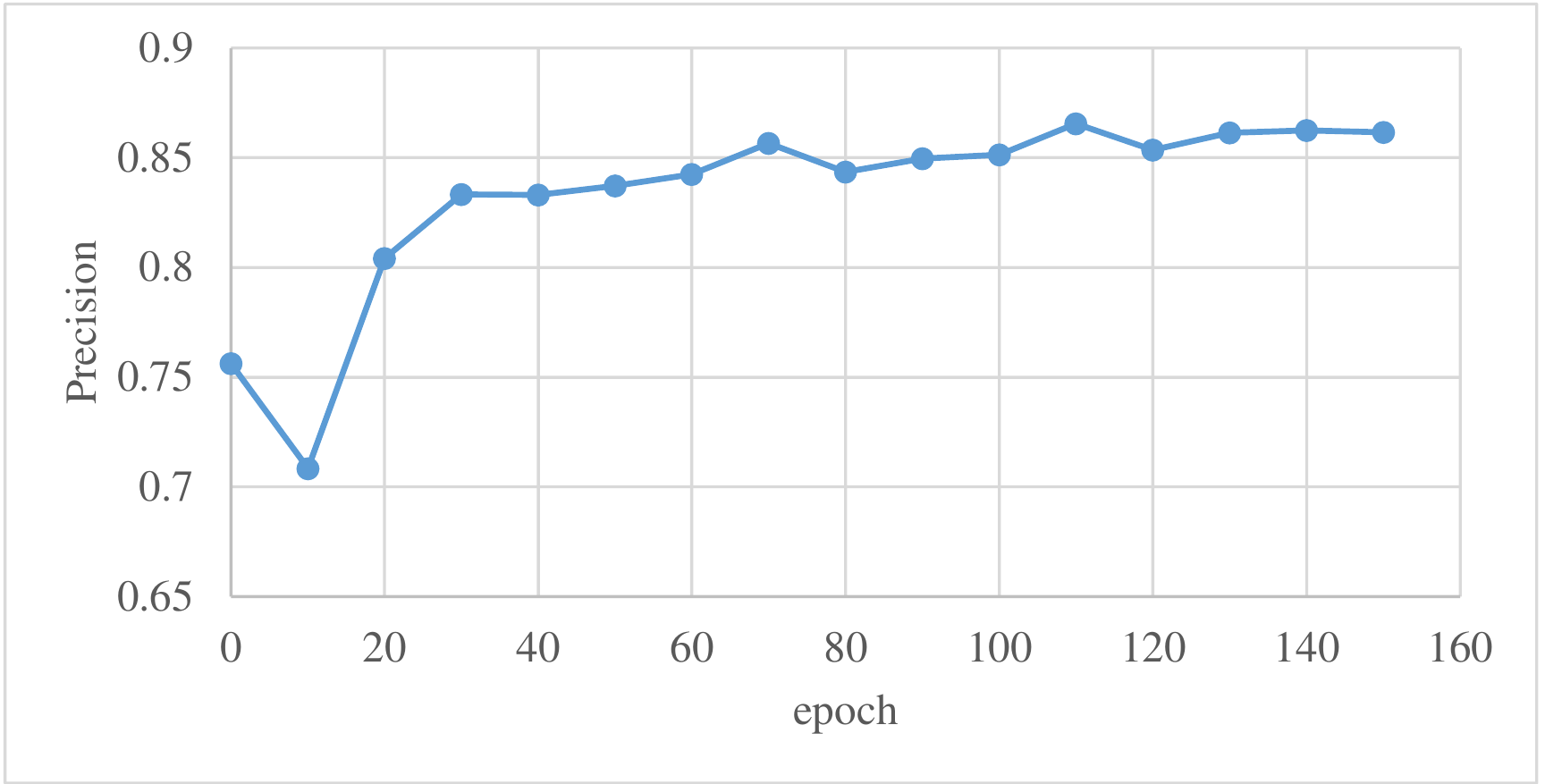}
    \caption{Precision for YOLOv5s SE.}
    \label{fig:mesh1}
\end{figure}
\begin{figure}[h]
    \centering
    \includegraphics[width=0.5\textwidth]{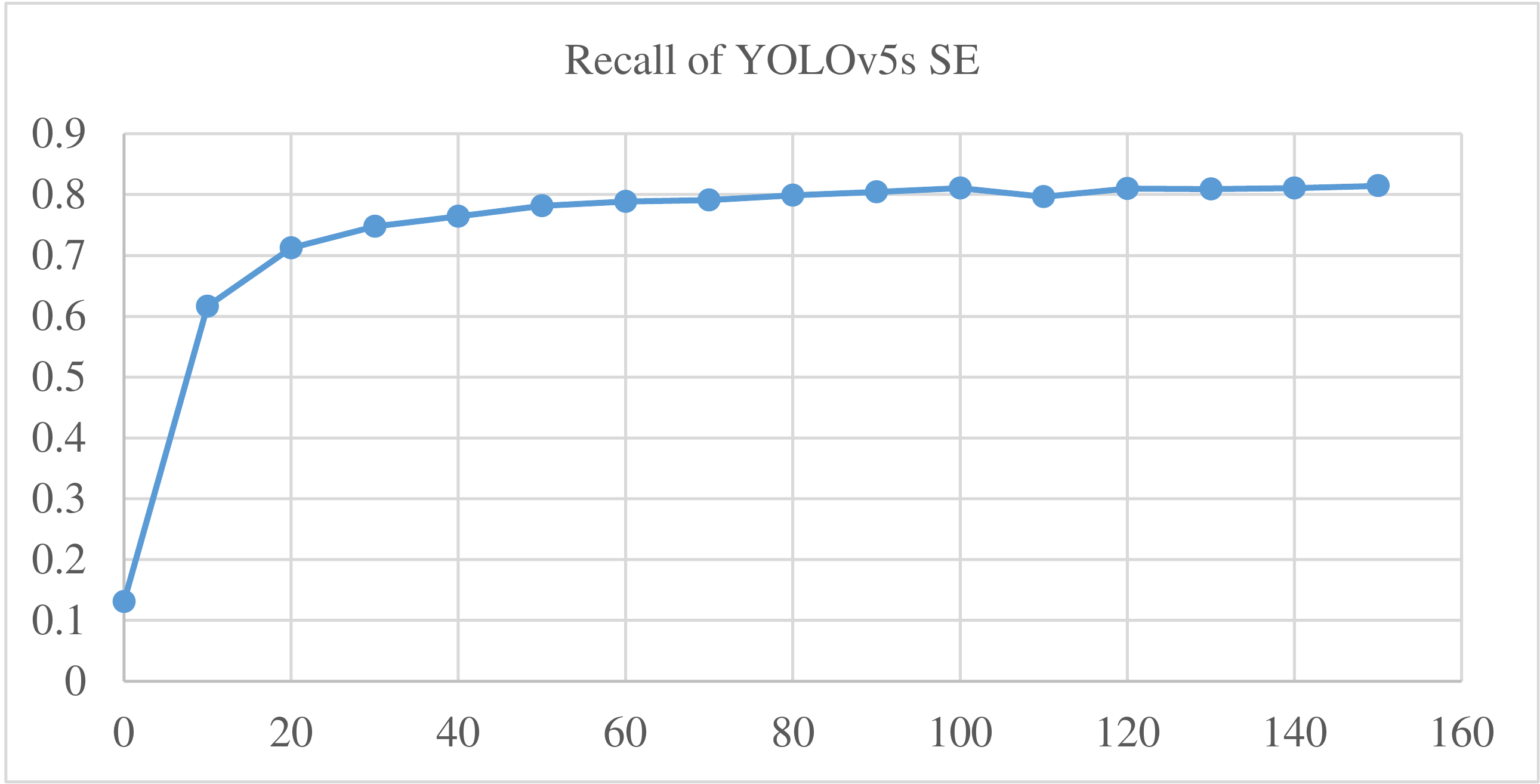}
    \caption{Recall for YOLOv5s SE.}
    \label{fig:mesh1}
\end{figure}
In addition, in order to have a comprehensive evaluation of Precision and Recall, the concepts of AP (Average Precision) and mAP (mean Average Precision) were introduced, trained on the YOLOv4 network framework using the same dataset and compared with the network used in this paper. It can be seen that this network improves the mAP by about 30$\%$ compared to YOLOv4, and has a higher advantage in the AP values of the subclasses of the stand, uphand. The results are shown in Figure 5.
\begin{equation}
    \centering
\begin{aligned}
AP_{i} = {\int_{0}^{1}{P(r)dr}}\\
mAP = \frac{1}{n}{\sum\limits_{i}^{n}\left( AP_{i} \right)}
\end{aligned}
\end{equation}
\begin{figure}[h]
    \centering
    \includegraphics[width=0.5\textwidth]{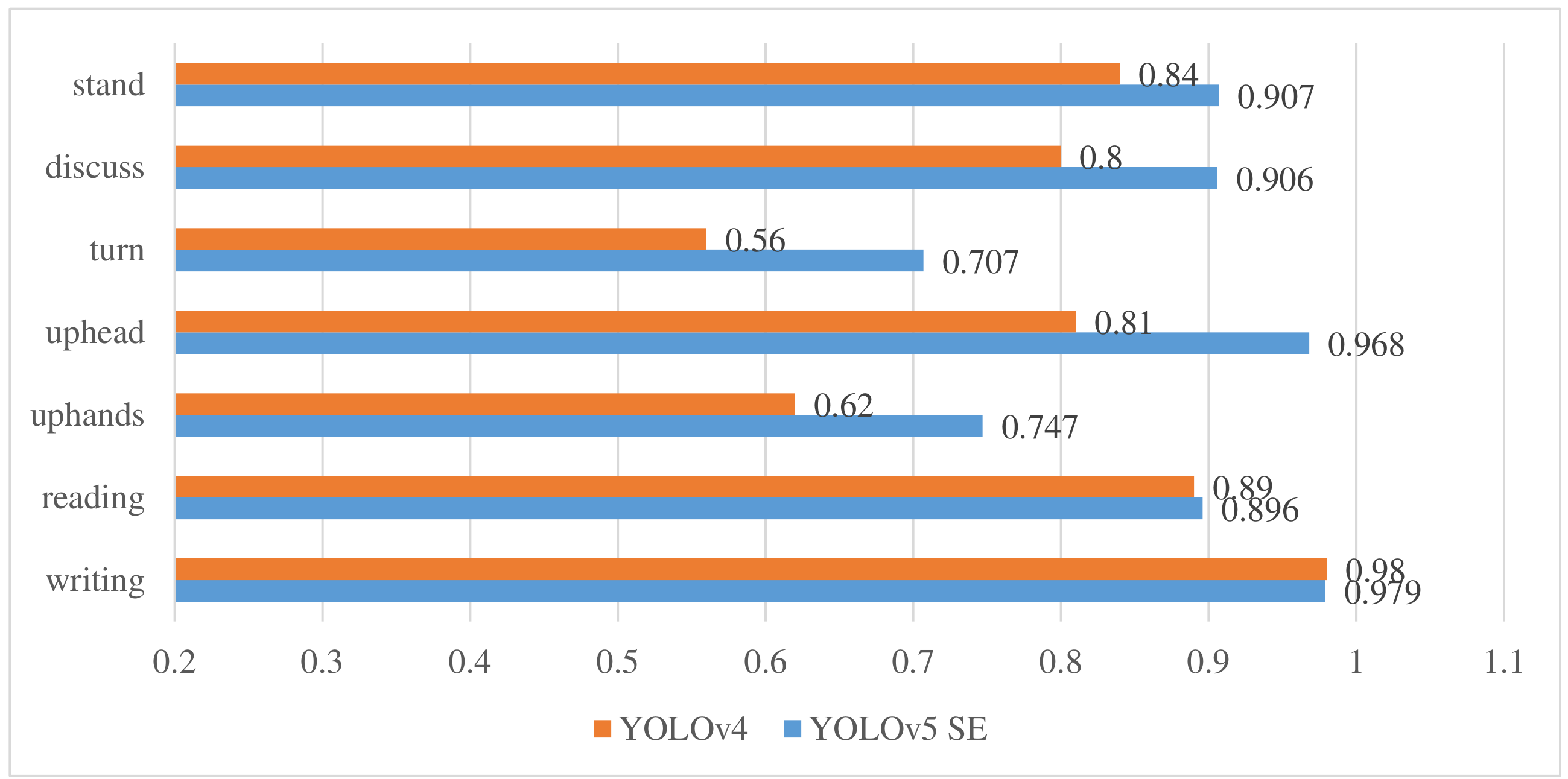}
    \caption{Results of this web comparison with YOLOv4.}
    \label{fig:mesh1}
\end{figure}
\begin{figure*}[h]
    \centering
    \includegraphics[width=1\textwidth]{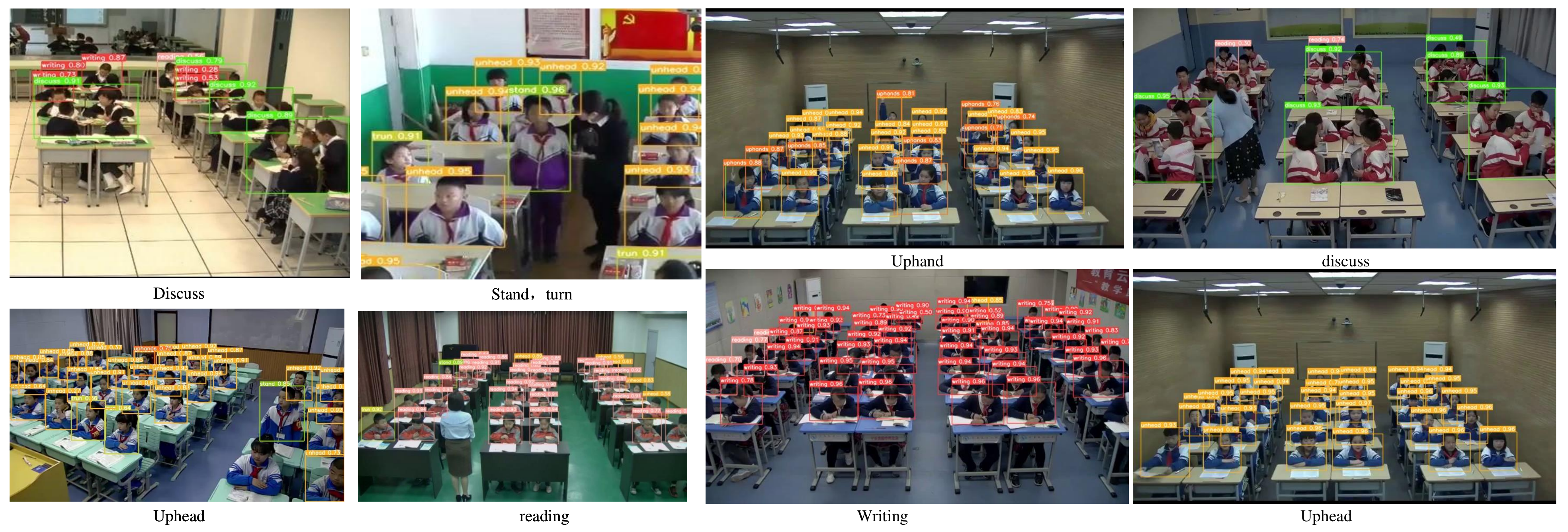}
    \caption{Student behavioral identification results.}
    \label{fig:mesh1}
\end{figure*}
\par Based on the Figure 6, it can be seen that the experiment was able to achieve relatively accurate recognition of students' group discussion, standing up, turning their heads, looking up to listen to the lecture, looking down to write and reading a book.
In addition, the same experiments were carried out in the same experimental setting for the YOLOv5l network with 53 convolutional layers and 3 pooling layers. The relevant mAP comparisons are as follows.\par
\begin{table}[]
    \centering
\begin{tabular}{ccc}
\multicolumn{3}{c}{Table 2. The mAP performance with different methods.}\\
\hline
epoch & yolov5l & yolov5s SE \\ \hline
0     & 0.59466 & 0.1491     \\
10    & 0.73182 & 0.69229    \\
30    & 0.83587 & 0.83085    \\
50    & 0.86475 & 0.85168    \\
70    & 0.87607 & 0.86676    \\
90& 0.88811 & 0.87308    \\
110   & 0.89193 & 0.87498    \\
130   & 0.89274 & 0.87936    \\
150   & 0.89283 & 0.88133    \\ \hline
\end{tabular}
\end{table}

\begin{figure}[h]
    \centering
    \includegraphics[width=0.5\textwidth]{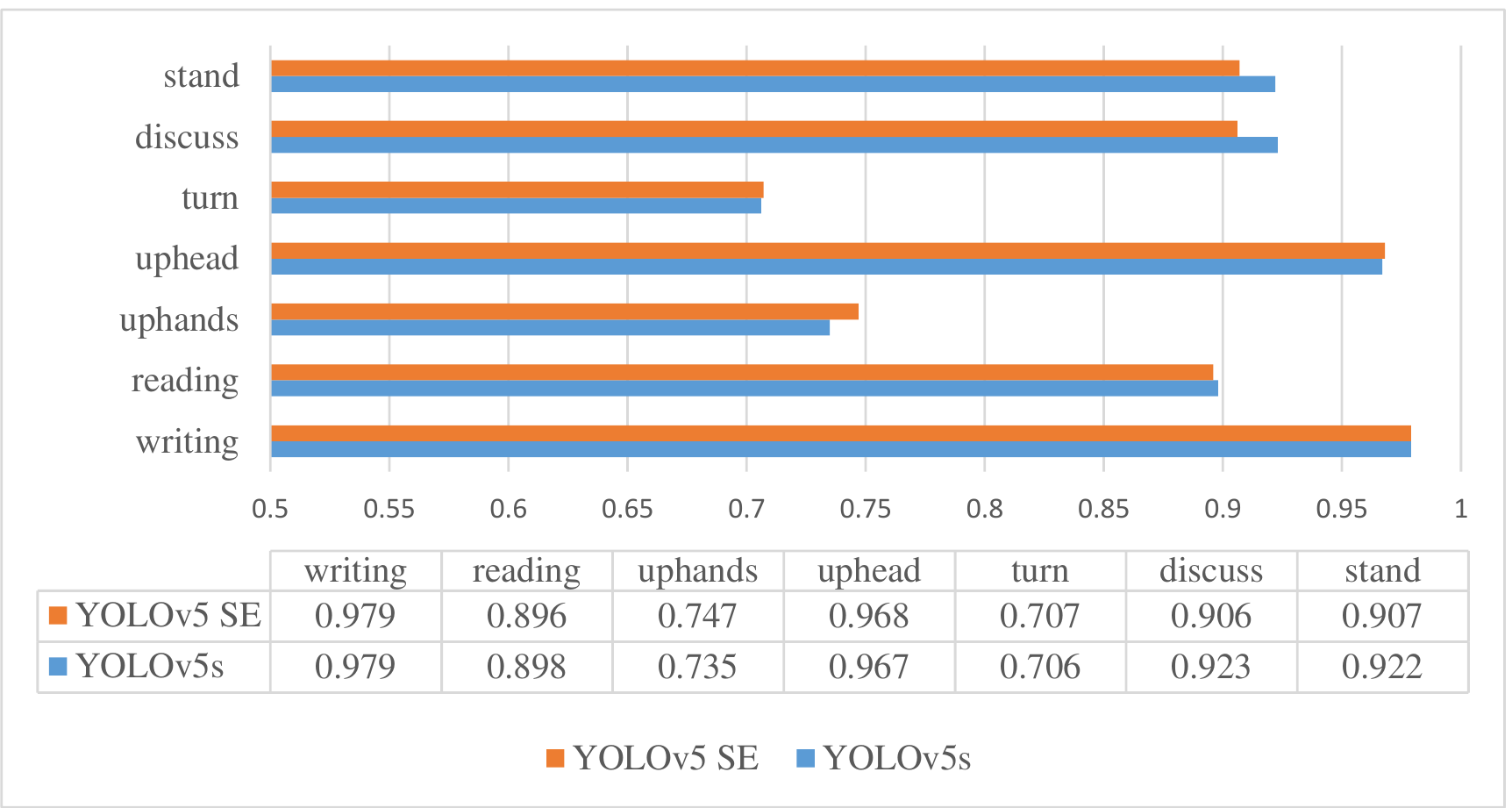}
    \caption{The AP for each class.}
    \label{fig:mesh1}
\end{figure}\begin{figure}[h]
    \centering
    \includegraphics[width=0.5\textwidth]{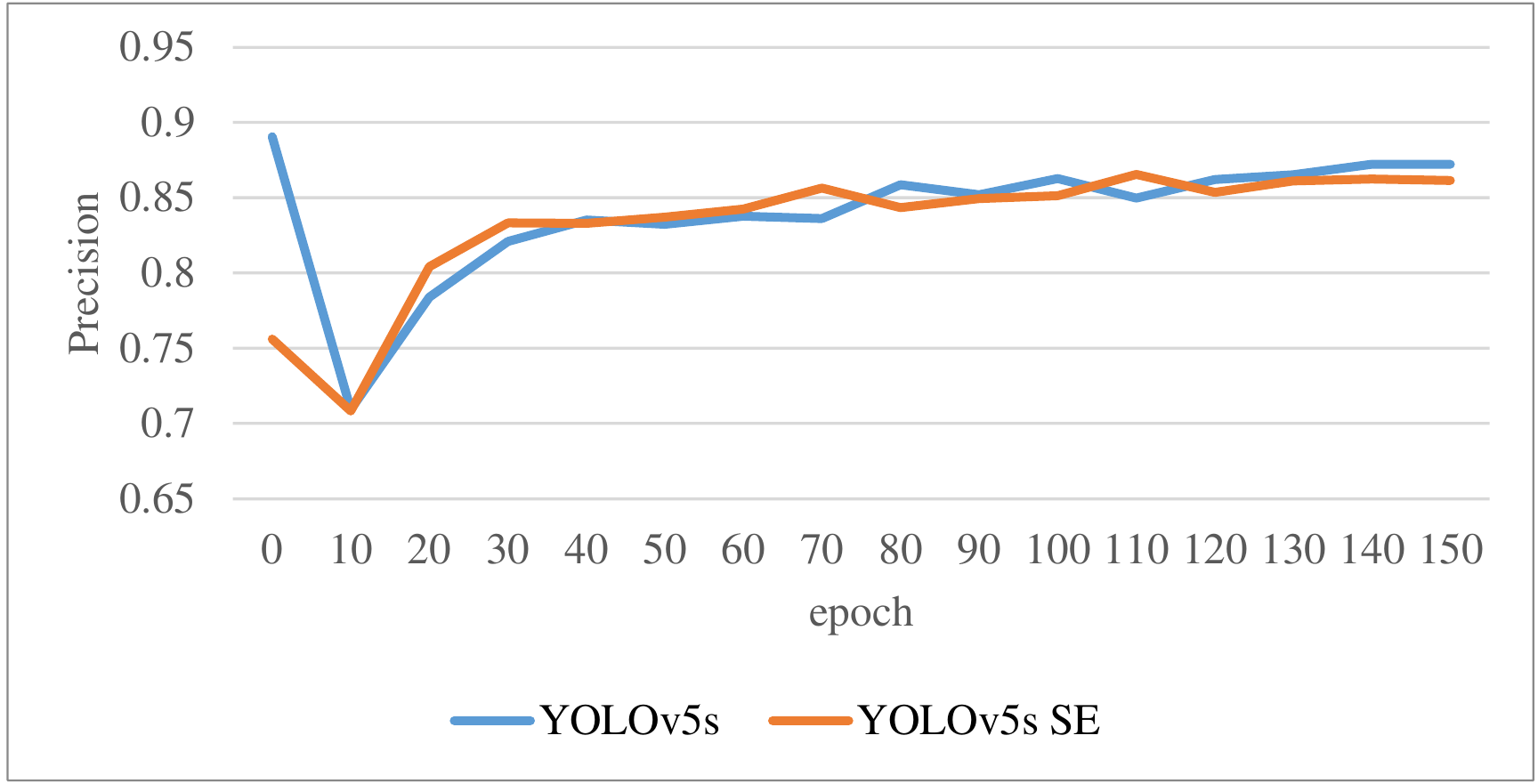}
    \caption{Precision of YOLOv5s and YOLOv5s SE.}
    \label{fig:mesh1}
\end{figure}\begin{figure}[h]
    \centering
    \includegraphics[width=0.5\textwidth]{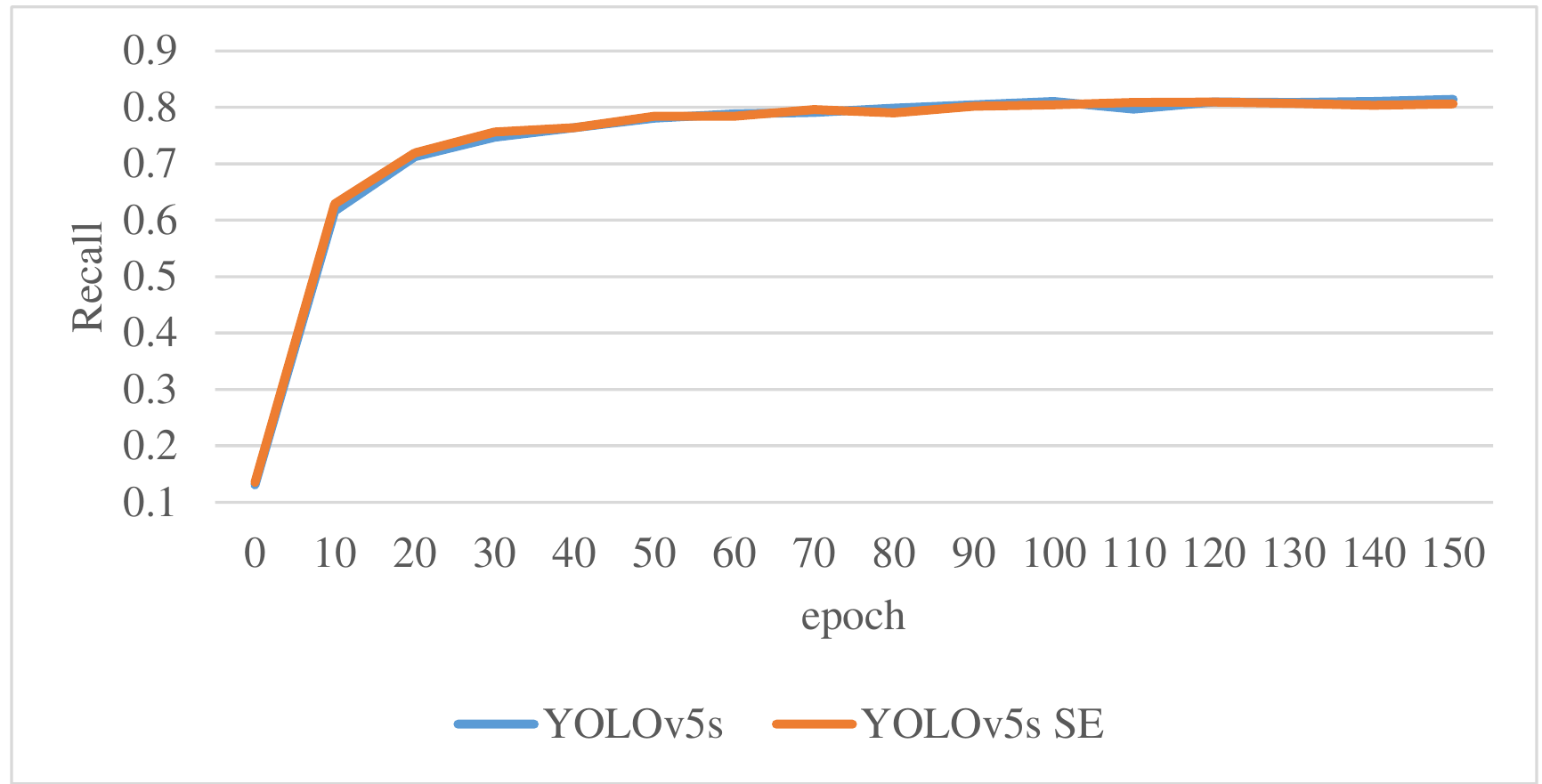}
    \caption{Recall of YOLOv5s and YOLOv5s SE.}
    \label{fig:mesh1}
\end{figure}

This paper also compares the performance of the YOLOv5s network without the introduction of the SE attention mechanism module. By comparing the AP values of each subclass and the precision and recall curves during the training process, it can be found that the detection results of the network with the SE mechanism added do not differ much from the original network. This may occur because the dataset used in this paper has a more homogeneous background and SE does not have a significant impact on the results.\par

It can be seen that there is little difference in performance between the network structure used in this paper and the YOLOv5l network, but the YOLOv5l network takes a long time to train and is not as fast as this network structure in terms of recognition speed. All things considered, the network used in this paper is more suitable for real-time detection in smart classrooms.\par

\section{Conclusion}
This experiment successfully made the recognition of common body postures appearing in students' classes, demonstrating the feasibility of using deep learning in smart classrooms. The performance of the YOLOv4, YOLOv5s and YOLOv5l networks was also compared, demonstrating the relative advantages of the networks in this paper. There are also some shortcomings in this experiment; the dataset selected was sourced from a model classroom, which did not incorporate some inappropriate student behavior in the classroom, and the dataset for this experiment was captured from the camera in front of the classroom, which can lead to false detection when performing recognition of side angles.

\section*{Acknowledgments}
The research work of this paper were supported by the National Natural Science Foundation of China (No. 62177022, 61901165, 61501199), Collaborative Innovation Center for Informatization and Balanced Development of K-12 Education by MOE and Hubei Province (No. xtzd2021-005), and Self-determined Research Funds of CCNU from the Colleges’ Basic Research and Operation of MOE (No. CCNU22QN013).

\bibliographystyle{ieeetr}
\bibliography{ref,Citations}

\end{document}